\documentclass[
 aps,
 prx,
 amsmath,amssymb,
 superscriptaddress,
 reprint
]{revtex4-2}

\usepackage{graphicx}
\usepackage{color}
\usepackage{dcolumn}
\usepackage{bm}

\usepackage{amsfonts}
\usepackage{subfigure}

\usepackage[utf8]{inputenc}
\usepackage[T1]{fontenc}
\usepackage{mathptmx}
\usepackage{etoolbox}

\usepackage{braket}
\usepackage{color}
\usepackage{algorithmic}
\usepackage{algorithm}
\usepackage{float} 

\usepackage[hidelinks]{hyperref}

\begin{document}

\title{
Training Energy-Based Models with Non-MCMC Samplers and Efficient Temperature Estimation
}

\author{Kentaro Kubo}
\email{kentaro.kubo.b91@mail.toshiba}
\affiliation{Corporate Laboratory, Toshiba Corporation, Kawasaki, Kanagawa 212-8582, Japan}
\author{Hayato Goto}
\affiliation{Corporate Laboratory, Toshiba Corporation, Kawasaki, Kanagawa 212-8582, Japan}
\affiliation{RIKEN Center for Quantum Computing (RQC), Wako, Saitama 351-0198, Japan}

\date{\today}

\begin{abstract} 
Efficient sampling from Boltzmann distributions over discrete variables is a fundamental operation in a wide range of applications.
While fast non-Markov chain Monte Carlo (MCMC) samplers have recently emerged as promising alternatives to conventional MCMC methods, their practical use for probabilistic learning remains hindered by the difficulty of estimating the effective temperature of the generated samples.
In this work, we begin by introducing Langevin simulated bifurcation (LSB), a novel Boltzmann sampler that enables fast and parallel sampling with accuracy comparable to sequential MCMC methods. 
To address the challenge of unknown effective temperature, we propose conditional expectation matching (CEM), an efficient estimation method applicable to energy-based models (EBMs) with exploitable conditional independence structures. 
Building on these components, we further develop a learning framework, termed sampler adaptive learning (SAL), which adaptively adjusts the model temperature to match that of the distribution induced by fast non-MCMC sampling.
We demonstrate the effectiveness of LSB, CEM, and SAL on semi-restricted Boltzmann machines (SRBMs), a class of EBMs that are difficult to train using conventional approaches. 
LSB achieves orders-of-magnitude acceleration over Gibbs sampling while maintaining comparable or higher accuracy, and CEM enables accurate temperature estimation of the resulting distribution with negligible computational overhead.
As a consequence, SAL enables efficient training of SRBMs and outperforms conventional Boltzmann machine learning methods on synthetic spin-glass datasets. 
In addition, the trained models achieve strong performance across multiple tasks, including image generation, reconstruction, and classification.
These results establish LSB as a fast and accurate Boltzmann sampler and provide key insights that enable practical applications of fast non-MCMC sampling methods via efficient temperature estimation with CEM.
\end{abstract}

\maketitle

\section{Introduction}
\begin{figure*}[htbp]
  \centering
  \includegraphics[width=\textwidth]{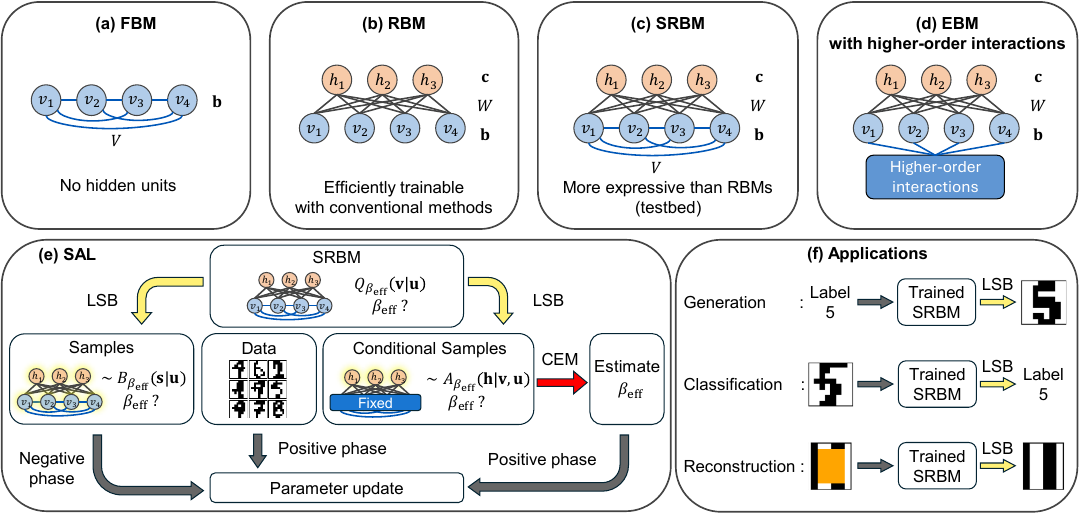}
  \caption{
  \textbf{Overview of model structures and proposed methods: }
    (a-d) Energy-based model structures: All models consist of visible nodes $v_i$ (blue circles, arranged in the lower row) and, if present, hidden nodes $h_j$ (orange circles, arranged in the upper row).
     (a) Fully-visible Boltzmann machine (FBM): only visible nodes are present, and all visible nodes are mutually connected.
  (b) Restricted Boltzmann machine (RBM): every visible node is connected to every hidden node, but there are no connections among visible nodes or among hidden nodes.
  (c) Semi-restricted Boltzmann machine (SRBM): obtained by adding pairwise connections among visible nodes to an RBM. 
  (d) Energy-based model (EBM) with higher-order interactions: obtained by extending SRBMs with higher-order interactions among visible units.
  (e)  Schematic of the SAL framework, which combines a non-MCMC sampler with conditional expectation matching (CEM, red arrow) for efficient estimation of the effective inverse temperature $\beta_{\rm eff}$. 
In this work, we use LSB (yellow arrow) as the sampler and train SRBMs.
  (f) Examples of applications of trained SRBMs: image generation, classification, and reconstruction.
}
  \label{fig:1}
\end{figure*}

Sampling from Boltzmann distributions over discrete variables is a fundamental computational primitive that appears across a wide range of scientific and engineering disciplines. 
A particularly prominent instance arises in machine learning, where energy-based models (EBMs)  rely on sampling to estimate gradients and generate data \cite{EBM1,EBM2}, with Boltzmann machines serving as a prominent example \cite{BML2,HintonCD,RBM_guide}. 
In quantum many-body physics, similar sampling tasks also arise in neural quantum states, including Boltzmann machine-based ans\"atze, where samples are used to evaluate observables and optimize variational wave functions \cite{BML_Quantum1}. 
More broadly, Boltzmann sampling plays a central role in a wide range of problems in statistical physics, Bayesian inference, and stochastic optimization \cite{Stat1,BayesianLearning,SA}, and its hardware implementations have also attracted attention as a potential pathway toward addressing the growing power consumption of modern AI systems \cite{pbit_review}.
Despite this broad applicability, efficient sampling from discrete Boltzmann distributions remains a major computational challenge. 
Conventional Markov chain Monte Carlo (MCMC) methods \cite{MCMC}, such as Gibbs sampling \cite{Gibbs} or the Metropolis–Hastings algorithm \cite{Metropolis,Hastings}, are inherently sequential and often exhibit poor scalability to practical large-scale applications. 
Furthermore, these methods often suffer from slow mixing in high-dimensional and rugged energy landscapes, severely limiting their practical efficiency. 
Therefore, there is a strong demand for alternative sampling approaches that can overcome these limitations.

In this work, we first introduce a novel Boltzmann sampler inspired by 
a quantum-inspired combinatorial optimization known as simulated bifurcation \cite{aSB,bSBdSB,HbSBHdSB,HigherorderSB,gSB}. 
We refer to this algorithm as Langevin simulated bifurcation (LSB). 
LSB enables parallel sampling while maintaining accuracy comparable to the conventional reliable method, namely sequential MCMC. 
However, a key challenge remains: the inverse temperature of the output distribution, $\beta_{\rm eff}$, is generally unknown and depends on both the problem instance and sampler parameters.
Importantly, this limitation is not unique to LSB, but is a common challenge encountered in other non-MCMC samplers, 
such as D-Wave's quantum annealers \cite{D-wave1,D-wave2,D-wave3,D-wave4,D-wave_Boltzmann,ex1,ex2,ex3}, 
coherent Ising machines \cite{CIM1,CIM2,CIM_Boltzmann}, 
their simulated counterparts \cite{SimCIM,SimCIM_Boltzmann}, 
quantum bifurcation machines \cite{KPO1}, 
noise-injected opto-electronic Ising machines \cite{noise}, 
networks of coupled optical parametric oscillators \cite{OPO}, 
and adaptive Ising machine \cite{AIM}. 
Unless $\beta_{\rm eff}$ is accurately estimated or controlled, 
sampling does not correspond to the desired distribution, leading to biased estimation of expectation values and probabilistic quantities, which in turn degrades downstream inference and learning tasks. 

However, the standard approach for estimating $\beta_{\rm eff}$, based on minimizing the Kullback–Leibler (KL) divergence, is computationally prohibitive within practical computational pipelines and becomes infeasible for large-scale problems \cite{beta_estimation}. 
Although alternative methods have been explored, they often suffer from accuracy and stability issues \cite{beta_estimation,QuALe,SimCIM_Boltzmann}. 
For the simplest class of EBMs, namely fully-visible Boltzmann machines (FBMs) [see Fig.\ \ref{fig:1}(a)], learning methods have been proposed to avoid explicit estimation of $\beta_{\rm eff}$ \cite{Gray-box}. 
As these models lack hidden variables, their expressive power is limited.
Thus, the use of fast non-MCMC samplers, including LSB, demands a scalable and accurate $\beta_{\rm eff}$ estimation method. 
Here, we introduce an efficient method for estimating $\beta_{\rm eff}$, which we term conditional expectation matching (CEM). 
The proposed method is applicable to EBMs in which conditional independence structures can be exploited. 
In particular, it is not limited to restricted Boltzmann machines (RBMs) [see Fig.\ \ref{fig:1}(b)], which can be efficiently trained using conventional methods \cite{HintonCD,RBM_guide}, but also extends to more expressive models such as semi-restricted Boltzmann machines (SRBMs) \cite{HintonSRBM,SRBM2,SRBM3} and their further extensions with higher-order interactions [see Fig.\ \ref{fig:1}(c) and (d)].
CEM is naturally parallelizable with sampling processes of non-MCMC samplers and provides accurate estimation of $\beta_{\rm eff}$. 
Furthermore, we propose a learning framework, termed sampler adaptive learning (SAL) [see Fig.\ \ref{fig:1}(e)], which combines fast non-MCMC sampling with CEM and adaptively adjusts the temperature of the model based on the distribution induced by the sampler.
This integration enables scalable and efficient training of EBMs with conditional independence structures.

As a representative testbed, we apply LSB, CEM, and SAL to SRBMs [see Fig.\ \ref{fig:1}(c)]. 
SRBMs extend RBMs by introducing interactions among visible units, thereby enhancing model expressiveness \cite{HintonSRBM,SRBM2,SRBM3}. 
However, efficient training of such models using conventional methods is challenging, as the introduction of these additional interactions breaks the conditional independence structure of the visible units exploited by block Gibbs sampling and contrastive divergence \cite{RBM_guide}. 
We first demonstrate that LSB enables fast and accurate sampling from the Boltzmann distribution of SRBMs, achieving accuracy comparable to or better than Gibbs sampling while providing orders-of-magnitude speedup. 
We further show that the inverse temperature of the resulting distribution can be efficiently and accurately estimated using CEM. 
These capabilities translate into improved learning performance, and we demonstrate that SAL-trained SRBMs outperform conventional learning methods in minimizing the training cost function on synthetic spin-glass datasets \cite{p-spin1,p-spin2}. 
Furthermore, we achieve good performance across multiple tasks, including high-fidelity image generation and reconstruction on the Bars-and-Stripes dataset \cite{BML3_Recognition,BAS2}, both unconditional and conditional image generation, and high classification accuracy on OptDigits \cite{OptDigits} [see Fig.\ \ref{fig:1}(f)].

The present results provide a proof of principle for the effectiveness of combining non-MCMC sampling and accurate temperature estimation via CEM, rather than establishing immediate competitiveness with state-of-the-art generative AI.
These ideas are expected to bridge fast non-MCMC sampling methods and their practical applications across a wide range of domains, including neural quantum states, more expressive EBMs, and energy-efficient hardware implementations of sampling-based learning.

\section{Energy-based model and Boltzmann distribution}
\label{sec:EBM}
Energy-based models (EBMs) are expressive probabilistic models that capture complex dependencies and latent structures \cite{EBM1,EBM2}.
We consider EBMs defined over visible units $\mathbf{v}$, which represent the observed data, and hidden units $\mathbf{h}$, which enhance the model's expressive capacity, forming an undirected model.
Both $\mathbf{v}$ and $\mathbf{h}$ are binary vectors of dimensions $N_v$ and $N_h$, respectively, taking values in $\{-1, +1\}$, like Ising spins.
The full state of the model is denoted by an $N$-dimensional Ising vector $\mathbf{s}=(\mathbf{v}^{\rm T},\mathbf{h}^{\rm T})^{\rm T}$, where $N=N_v+N_h$.
In generative modeling, the EBM often models a target distribution $P_{D}(\mathbf{v})$ using a marginal Boltzmann distribution 
\begin{align}
	Q_{\beta}(\mathbf{v}|\mathbf{u})=\sum_{\mathbf{h}}B_{\beta}(\mathbf{s}|\mathbf{u}),
\end{align}
where 
\begin{align}
	B_{\beta}(\mathbf{s}|\mathbf{u})=\frac{e^{-\beta E(\mathbf{s}|\mathbf{u})}}{Z_{\beta}}
\end{align}
is a Boltzmann distribution with inverse temperature $\beta$, $Z_{\beta}=\sum_{\mathbf{s}}e^{-\beta E(\mathbf{s}|\mathbf{u})}$ is the partition function, $E(\mathbf{s}|\mathbf{u})$ is the global energy function, $\mathbf{u}$ denotes the model parameter. 
Throughout this paper, for an Ising vector $\mathbf{S}$, the notation $\sum_{\mathbf{S}}$ denotes the summation over all configurations of $\mathbf{S}$.
Learning aims to minimize the Kullback-Leibler (KL) divergence 
\begin{align}
	D_{\rm KL}(P_D||Q_\beta)=\sum_{\mathbf{v}}P_D(\mathbf{v})\log\frac{P_D(\mathbf{v})}{Q_{\beta}(\mathbf{v}|\mathbf{u})}
	\label{eq:DKL}
\end{align}
for $\mathbf{u}$, typically via gradient-based methods.
Since the exact calculation of the gradients scales exponentially with the system size, it is standard to approximate the gradients using Boltzmann sampling, most commonly via MCMC techniques such as Gibbs sampling \cite{Gibbs} or the Metropolis–Hastings algorithm \cite{Metropolis,Hastings}.
However, these MCMC methods are often computationally inefficient due to their inherent sequentiality and slow mixing property, motivating the need for fast and accurate alternative sampling approaches.

Among the above EBMs, those whose energy function takes the quadratic form
\begin{align}
	E(\mathbf{s}|\mathbf{u})=-\frac{1}{2}\mathbf{s}^{\rm T}J\mathbf{s}-\mathbf{f}\cdot\mathbf{s}. 
\end{align}
are referred to in particular as Boltzmann machines.
Here, $J$ is an $(N \times N)$ weight matrix satisfying $J_{ii}=0$ and $J_{ij}=J_{ji}\in\mathbb{R}$ for $i \neq j$, and $\mathbf{f}$ is an $N$-dimensional bias vector with $f_i \in \mathbb{R}$. 
The set of model parameters is denoted by $\mathbf{u}=(J,\mathbf{f})$.
For specific architectures, the parameter set $\mathbf{u}$ for Boltzmann machines can be expressed according to their structure. 
In particular, for fully-visible, restricted, and semi-restricted Boltzmann machines (FBMs, RBMs, SRBMs), the parameters can be written as $\mathbf{u} = (V, \mathbf{b})$, $(W, \mathbf{b}, \mathbf{c})$, and $(V, W, \mathbf{b}, \mathbf{c})$, respectively, 
where $V$ and $W$ denote the weight matrices between visible--visible and visible--hidden units, respectively, and $\mathbf{b}$ and $\mathbf{c}$ are bias vectors for the visible and hidden units, respectively. 

In the following section, we focus on SRBMs \cite{HintonSRBM,SRBM2,SRBM3} as a representative testbed for our approaches (LSB, CEM, and SAL), whose energy function is given by
\begin{align}
	E(\mathbf{s}|\mathbf{u})=
	-\frac{1}{2}\mathbf{v}^{\rm T}V\mathbf{v}-\mathbf{v}^{\rm T}W\mathbf{h}
	-\mathbf{b}\cdot\mathbf{v}
	-\mathbf{c}\cdot\mathbf{h}.
	\label{eq:SRBM}
\end{align}
Because of the additional interactions among visible units, 
SRBMs are more expressive than the celebrated RBMs \cite{HintonCD,RBM_guide}, but their efficient training remains challenging using conventional methods due to the lack of available parallel sampling.
As will be shown in Sec.\ \ref{sec:LSB}, the proposed LSB overcomes this limitation by enabling parallel sampling while maintaining high accuracy.
In addition, the proposed CEM allows accurate estimation of the effective inverse temperature associated with the distribution induced by the sampler, as will be demonstrated in Sec.\ \ref{sec:CEM}.
By combining these components, the SAL framework enables effective training of SRBMs, achieving performance beyond conventional Boltzmann machine learning approaches (see Sec.\ \ref{sec:Experiments}).
While SRBMs serve as a convenient testbed in this study, we note that the proposed framework can be applied to EBMs with exploitable conditional independence structures.

\section{Langevin simulated bifurcation (LSB)}
\label{sec:LSB}
Several optimization machines and algorithms, such as D-Wave's quantum annealers, 
coherent Ising machines,  
their simulated counterparts, 
quantum bifurcation machines, 
noise-injected opto-electronic Ising machines,
networks of coupled optical parametric oscillators, 
and adaptive Ising machine
have been shown to function effectively as Boltzmann samplers \cite{D-wave1,D-wave2,D-wave3,D-wave4,D-wave_Boltzmann,ex1,ex2,ex3,CIM1,CIM2,CIM_Boltzmann,SimCIM,SimCIM_Boltzmann,noise,KPO1,OPO,AIM}. 
Inspired by these findings, we investigated the output distributions of another class of optimization algorithms, simulated bifurcation algorithms \cite{aSB,bSBdSB,HbSBHdSB,HigherorderSB,gSB}.
Our results show that special cases of conventional simulated bifurcation algorithms approximate the Boltzmann distribution to a nontrivial degree, but their sampling accuracy is substantially lower than that achieved by MCMC methods (see Appendix\ \ref{sec:SK}).
In this section, we propose a modified sampling algorithm that achieves sampling accuracy comparable to MCMC.
We refer to this algorithm as Langevin Simulated Bifurcation (LSB), since its continuous limit resembles the celebrated continuous-variable sampler, Langevin Monte Carlo, whose stationary distribution is guaranteed to be the Boltzmann distribution \cite{LMC1,LMC2}.

Similarly to discrete simulated bifurcation (dSB) \cite{bSBdSB}, LSB is based on the following equations of motion for positions $x_i$ and momenta $y_i$ corresponding to $s_i$ in the general global energy function $E(\mathbf{s} \mid \mathbf{u})$: 

\begin{align}
	\dot{y}_i &= -\left. \frac{\partial E(\mathbf{x}|\mathbf{u})}{\partial x_i} \right|_{\mathbf{x}=\mathbf{s}}, 
	\label{eq:ydot}
	\\
	\dot{x}_i &= y_i,
	\label{eq:xdot}
\end{align}
where the dots denote time derivatives.
Equations (\ref{eq:ydot}) and (\ref{eq:xdot}) are solved with the symplectic Euler method \cite{SymplecticEuler}: 
\begin{align}
	y_i(k+1)&=y_i(k)
	-\left. \frac{\partial E[\mathbf{x}(k)|\mathbf{u}]}{\partial x_i} \right|_{\mathbf{x}=\mathbf{s}}\Delta, \\
	x_i(k+1)&=x_i(k)+y_i(k+1)\Delta,
\end{align}
where $\Delta$ is the time step and
$k\in\{0,\cdots,M\}$ denotes iteration. 
To improve sampling accuracy, at the end of each iteration, LSB applies discretization and stochastic initialization to $x_i$ and $y_i$, respectively:
\begin{align}
	&x_i \rightarrow {\rm sgn} (x_i), 
	\label{eq:discretization}
	\\
	&y_i \rightarrow \xi_i \sim \mathcal{N}(0,\sigma), 
\end{align}
where 
$\xi_i$ is a stochastic variable drawn from a Gaussian distribution $\mathcal{N}(\mu,\sigma)$ with mean $\mu=0$ and standard deviation $\sigma$, 
and the notation $\sim$ indicates that the variable on the left-hand side follows the distribution on the right-hand side.
Here, $\Delta$ and $\sigma$ are the only hyperparameters of LSB. 
Note that the update rule inherits the properties of the conventional simulated bifurcation algorithm and can be implemented in parallel for each component. 
In this work, $x_i$ and $y_i$ are initialized by sampling uniformly from $\{-1, 1\}$ and from $N(0,\sigma)$ using the same $\sigma$ as in the update rule, respectively, though other choices are possible in general.  
After $M$ iterations of the above updates, the resulting $\mathbf{x}$ provides a sample of $\mathbf{s}={\rm sgn}(\mathbf{x})$; 
executing this procedure $L$ times, we obtain a sampled Ising vector set $\{\mathbf{s}^{(l)}|l=1,\cdots,L\}$.

To assess the sampling performance of LSB, we evaluated its accuracy against conventional Gibbs sampling (MCMC) at $\beta=1$ on random SRBMs with $N_v=10$ and $N_h=5$, 
using $M=100$ iterations for LSB and the same $M$ Monte Carlo steps for Gibbs sampling. 
Edge weight $V_{ij}$ and $W_{ij}$ were sampled from $\mathcal{N}\left(0, 2/\sqrt{N}\right)$ with $N=N_v+N_h=15$, and bias $b_i$ and $c_i$ were set to zero. 
Sampling accuracy was defined as the KL divergence $D_{\mathrm{KL}}(P_S \| B_{\beta_{\rm eff}})$, where
$P_S(\mathbf{s})=\frac{1}{L}\sum_{l=1}^{L}\delta(\mathbf{s}-\mathbf{s}^{(l)})$ 
represents the empirical distribution of the sampled Ising vectors, 
and $\delta(\mathbf{x})=1$ if $\mathbf{x}=\mathbf{0}$ and 0 otherwise.
The effective inverse temperature $\beta_{\rm eff}$ was determined self-consistently 
by minimizing the KL divergence \cite{beta_estimation}: $\beta_{\rm eff}={\rm argmin}_\beta D_{\mathrm{KL}}(P_S \| B_{\beta})$. 
We initialized $\beta$ at 1 and performed the minimization over $[0,\infty)$ using the L-BFGS-B method implemented in \texttt{scipy.optimize.minimize} \cite{scipy}.

Figure \ref{fig:LSB_accuracy_CEM} shows the sampling accuracy $D_{\mathrm{KL}}(P_S \| B_{\beta_{\rm eff}})$ and the corresponding $\beta_{\rm eff}$ values for 10 random instances. 
We fixed $\Delta$ at 1 and optimized $\sigma$ from the candidate set $\sigma^{-2}\in\{0.5, 0.6, \cdots, 1.9, 2.0\}$ to maximize LSB sampling accuracy for each instance.
LSB outperformed Gibbs sampling in 6 out of 10 instances based on the sampling accuracy metric $D_{\mathrm{KL}}(P_S || B_{\beta_{\rm eff}})$. 
Across all 10 instances, the mean $D_{\mathrm{KL}}(P_S || B_{\beta_{\rm eff}})$ ($\pm$ standard error) was $0.09\pm0.02$ for Gibbs and $0.07 \pm 0.01$ for LSB.
Taking the errors into account, LSB demonstrates performance comparable to {or even better than Gibbs sampling. 
We use LSB for SAL in Sec.\ \ref{sec:Experiments}.

We also confirmed that the same qualitative trends persist for more complex Boltzmann machines such as the Sherrington–Kirkpatrick  model (see Appendix~\ref{sec:SK}), as well as 
a system size 20\% larger than the original setting (see Appendix~\ref{sec:additional_computational}).
In addition, as discussed in Appendix~\ref{sec:additional_computational},
elapsed-time measurements on CPU under the same computational resources
show that LSB achieves these improvements while providing orders-of-magnitude speedup over Gibbs sampling, 
with memory usage of the same order of magnitude.
This speed advantage reflects the intrinsically sequential nature
of Gibbs updates, in contrast to the matrix--vector operations underlying LSB,
which are naturally amenable to parallel execution and are therefore
expected to benefit further from hardware acceleration such as FPGA implementations.

Furthermore, to assess the practical role of the discretization and
stochastic-initialization steps introduced in LSB,
we performed additional comparisons with closely related variants,
as detailed in Appendix~\ref{sec:SK}.
Although a complete theoretical explanation of the observed behavior
remains open, these comparisons provide clear empirical evidence that
both discretization and stochastic initialization are important for
achieving improved sampling accuracy in LSB.
\begin{figure}[t]
	\centering
	\includegraphics[width=0.5\textwidth]{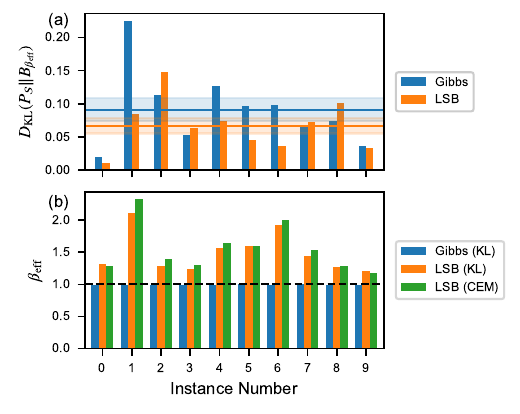}
	\caption{
	\textbf{Sampling accuracy and effective inverse temperature for LSB and Gibbs sampling:}
	(a) Sampling accuracy $D_{\mathrm{KL}}(P_S||B_{\beta_{\rm eff}})$ for LSB and Gibbs sampling across 10 random instances of SRBM with $N_v=10$ and $N_h=5$. 
	For each instance, $\Delta$ was fixed at 1 and $\sigma$ was optimized from the candidate set $\sigma^{-2}\in\{0.5, 0.6, \dots, 2.0\}$ to maximize LSB sampling accuracy. 
	Horizontal lines and shaded regions indicate the means KL divergence and their standard errors: $0.09 \pm 0.02$ for Gibbs and $0.07 \pm 0.01$ for LSB.
	(b) The effective inverse temperature $\beta_{\rm eff}$ for the output distribution produced by each sampler. 
	For Gibbs sampling, $\beta_{\rm eff}$ was estimated by minimizing the KL divergence.
	For LSB, $\beta_{\rm eff}$ was estimated by both minimizing the KL divergence and CEM. 
	}
	\label{fig:LSB_accuracy_CEM}
\end{figure}

As noted above, LSB resembles Langevin Monte Carlo in the continuous-time limit, whose stationary distribution is known to follow a Boltzmann distribution \cite{LMC1,LMC2}.
Therefore, LSB is also expected to generate samples that approximately follow a Boltzmann distribution.
However, for LSB, there is currently no theoretical guarantee that the generated distribution is exactly described by a Boltzmann distribution.
Moreover, even when the output distribution approximates a Boltzmann distribution, its effective inverse temperature generally depends on the problem instance and sampler parameters due to discretization effects.
As a result, the effective inverse temperature cannot be determined in advance.
Therefore, estimating the effective temperature is essential for practical applications of LSB.
In the next section, we introduce an efficient method for estimating this effective temperature.

\section{Conditional expectation matching (CEM)}
\label{sec:CEM}
Figure \ref{fig:LSB_accuracy_CEM}(b) shows that the effective inverse temperature $\beta_{\rm eff}$ estimated by 
minimizing the KL divergence varies across instances for LSB sampling (orange), whereas Gibbs sampling consistently yields $\beta_{\rm eff} \simeq 1$ (blue). 
Estimating the unknown $\beta_{\rm eff}$ is essential for successful probabilistic learning.
However, the conventional approach based on minimizing the KL divergence \cite{beta_estimation}
scales exponentially with the system size $N$, making it intractable for practical problems.
Thus, the instance-dependent variability of $\beta_{\rm eff}$ in LSB output distributions poses a significant challenge to Boltzmann machine learning with LSB. 
To overcome this, we propose a new and efficient $\beta_{\rm eff}$ estimation method: conditional expectation matching (CEM), as detailed below.

In a Boltzmann distribution $B_{\beta_{\rm eff}}(\mathbf{s}|\mathbf{u})$ of an SRBM, 
hidden variables are conditionally independent when the visible variables $\mathbf{v}$ are fixed to a condition vector $\mathbf{r}$. 
This property allows analytical expressions for their conditional expectations \cite{RBM_guide}:  
\begin{align}
	\langle h_j \rangle_{\mathbf{r},\beta_{\rm eff}}
	&:=\sum_{\mathbf{h}} h_j A_{\beta_{\rm eff}}(\mathbf{h}|\mathbf{v},\mathbf{u})\delta(\mathbf{v}-\mathbf{r}), \nonumber\\
	&= \tanh \left [ \beta_{\rm eff}\left(c_{j}+\sum_{i=1}^{N_v}r_i W_{ij}\right)\right ],
	\label{eq:hi}
\end{align}
where
\begin{align}
	A_{\beta_{\rm eff}}(\mathbf{h}|\mathbf{v},\mathbf{u})
	&:=
	\frac{B_{\beta_{\rm eff}}(\mathbf{s}|\mathbf{u})}
		{\sum_{\mathbf{h}}B_{\beta_{\rm eff}}(\mathbf{s}|\mathbf{u})},
\end{align}
is a conditional Boltzmann distribution. 
On the other hand, $\langle h_j \rangle_{\mathbf{r},\beta_{\rm eff}}$ can also be directly obtained by conditional LSB sampling (see Appendix\ \ref{sec:ConditionalSampling}).  
Hereafter, 
this sampling-based estimation of $\langle h_j \rangle_{\mathbf{r},\beta_{\rm eff}}$ is denoted by  $\langle h_j \rangle_{\mathbf{r},C}$. 
We estimate $\beta_{\rm eff}$ by minimizing the squared difference between $\langle h_j \rangle_{\mathbf{r},C}$ and the right-hand side of Eq.\ (\ref{eq:hi})
across all hidden units:
\begin{align}
	&\beta_{\rm eff}={\rm argmin}_{\beta>0} F_{\mathbf{r}}(\beta), \\
	&F_{\mathbf{r}}(\beta):=\sum_{j=1}^{N_h}\left\{ \langle h_j \rangle_{\mathbf{r},C}-\tanh \left[\beta\left(c_{j}+\sum_{i=1}^{N_v}r_i W_{ij}\right)\right]\right\}^2.
\end{align}
The minimization is performed similarly to the above 
minimization of the KL divergence.
This procedure, termed conditional expectation matching (CEM), leverages consistency between conditionally sampled expectations $\langle h_j \rangle_{\mathbf{r},C}$
and analytical conditional expectations expressed as a function of $\beta_{\rm eff}$ [the right-hand side of Eq.\ (\ref{eq:hi})].
CEM can be executed in parallel with full-system sampling to approximate $B_{\beta_{\rm eff}}(\mathbf{s}|\mathbf{u})$, in sharp contrast to conventional $\beta_{\rm eff}$ estimators \cite{beta_estimation, CIM_Boltzmann, QuALe, SimCIM_Boltzmann}. 
Furthermore, the conditional sampling cost to compute $\langle h_j \rangle_{\mathbf{r},C}$ is lower than full-system sampling. 
Therefore, CEM does not become a bottleneck during learning (see also Appendix~\ref{sec:additional_computational}).
The condition vector $\mathbf{r}$ can be randomly generated or selected from the dataset during learning, but is not restricted to these choices.

In Fig.\ \ref{fig:LSB_accuracy_CEM}(b),
we compare $\beta_{\rm eff}$ estimated by CEM (green) with that obtained by minimizing the KL divergence (orange) for LSB output distributions.
For CEM, each component of the condition vectors $\mathbf{r}$ was uniformly sampled from $\{-1, 1\}$.
The mean signed relative error between them across 10 instances is approximately 3.6\%.
Comparisons with other inverse-temperature estimation methods, including modified versions of CEM (CEM-$n$), are provided in the Supplemental Material~\cite{sp}.

Thus, CEM enables accurate inverse temperature estimation. 
Moreover, because it is parallelizable with LSB sampling, it does not become a bottleneck in training,  unlike 
estimation based on minimizing the KL divergence \cite{beta_estimation}.
By leveraging inverse temperature estimation through CEM in learning, 
we overcome the computational difficulty of the positive phase (see Appendix\ \ref{sec:DetailsOfSAL} and Ref.\ [\onlinecite{Gray-box}]).

While the present demonstration is performed using LSB as the sampler, CEM is not tied to this particular choice.
In principle, similar performance is expected to be possible with other non-MCMC samplers.
Moreover, since CEM relies only on the conditional independence of hidden variables, it is not restricted to the SRBM setting studied here.
As long as this structural property [Eq.\ (\ref{eq:hi})] is satisfied, CEM is expected to be applicable to EBMs beyond SRBMs.
For example, one could consider extensions to models with higher-order interactions in the visible layer [see Fig.\ \ref{fig:1}(d)], beyond the standard pairwise couplings in Boltzmann machines.

\section{Sampler adaptive learning (SAL)}
\label{sec:Experiments}
In conventional energy-based generative learning [see Sec.\ \ref{sec:EBM}], the inverse temperature $\beta$ of the model distribution $Q_{\beta}(\mathbf{v}|\mathbf{u})$ is fixed, typically to 1. 
In contrast, we introduce a learning framework, termed sampler adaptive learning (SAL), in which
$\beta$ is set to $\beta_{\rm eff}$, the inverse temperature of the output distribution produced by a sampler, which can depend on both the model parameters $\mathbf{u}$ and the sampling settings (see Sec.~\ref{sec:CEM}).
As $\mathbf{u}$ is updated during learning, $\beta_{\rm eff}$ also changes. 
Thus, SAL involves not only updating the learning parameters $\mathbf{u}$ but also adapting the model function $Q_{\beta_{\rm eff}}$ to the inverse temperature $\beta_{\rm eff}$. 
This stands in sharp contrast to the conventional learning procedure, which updates only $\mathbf{u}$ while keeping $\beta$ (and consequently the model function $Q_{\beta}$) fixed. 
Learning is performed by gradient descent on $D_{\rm KL}(P_{D}||Q_{\beta_{\rm eff}})$.
The gradient decomposes into a negative phase, defined as an expectation under the Boltzmann distribution $B_{\beta_{\rm eff}}(\mathbf{s}|\mathbf{u})$, and a positive phase, defined using the data distribution $P_{D}$ together with $\beta_{\rm eff}$ [see Appendix\ \ref{sec:DetailsOfSAL} for the details].
In this work, we adopt LSB as a sampler and CEM as an efficient $\beta_{\rm eff}$ estimator.

In the remainder of this section, we apply SAL to train SRBMs on three distinct datasets, 
to demonstrate its scalability and effectiveness. 
The target distributions $P_D(\mathbf{v})$ are set as the empirical distribution of each dataset $\{\mathbf{v}^{(d)} | d = 1,\dots,D\}$, where each data point $\mathbf{v}^{(d)}$ is an $N_v$-dimensional Ising vector: $P_D(\mathbf{v}) = \frac{1}{D} \sum_{d=1}^D \delta(\mathbf{v} - \mathbf{v}^{(d)})$.
Details of each dataset and the training settings are provided in the Supplemental Material~\cite{sp}.

\subsection{Learning performance on the 3-spin model}
\label{sec:3spinmodel}
\begin{figure}[t]
	\includegraphics[width=0.48\textwidth]{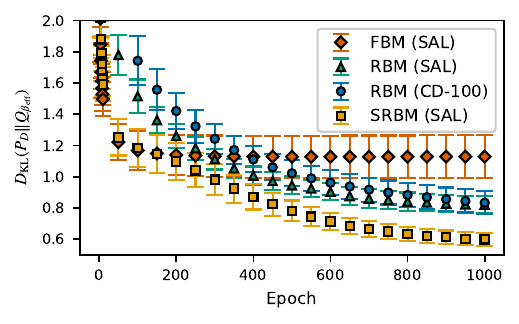}
	\caption{
	\textbf{Learning performance on the 3-spin model:} 
	The vertical axis shows the cost function $D_{\rm KL}(P_D||Q_\beta)$, and the horizontal axis indicates training epochs. 
	Four models are compared: 
	FBMs, RBMs, and SRBMs trained using SAL;  
	RBMs with $\beta = 1$ trained using CD-100. 
	The number of hidden variables $N_h$ for both RBMs and SRBMs was set to 5. 
	Each point represents the mean over 10 independent instances, and error bars denote the standard error. 
	Lower $D_{\rm KL}(P_D||Q_\beta)$ values indicate closer agreement between $Q_\beta$ and $P_D$. 
	}
	\label{fig:3spin}
\end{figure}
\begin{figure*}[t]
	\includegraphics[width=1.0\textwidth]{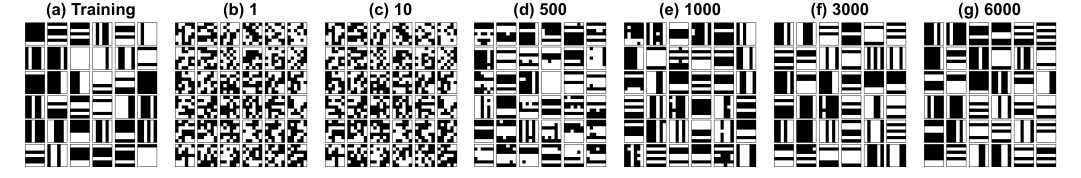}
	\caption{
	\textbf{Generative performance on the BAS dataset:}
	(a) 36 randomly selected training samples.
(b)–(g) 36 samples generated by the trained SRBM with LSB after 1, 10,  500, 1000, 3000, and 6000 training epochs, respectively.
	}
	\label{fig:BAS_Generation}
\end{figure*}
\begin{figure*}[t]
	\includegraphics[width=0.965\textwidth]{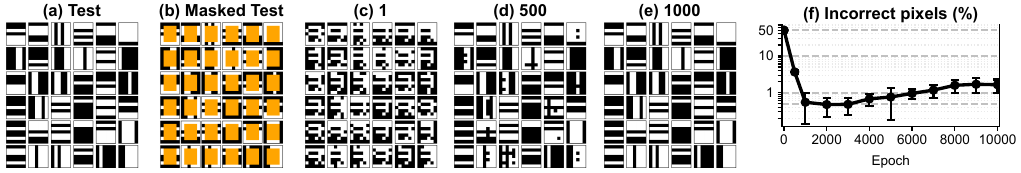}
	\caption{
	\textbf{Reconstruction performance on the BAS dataset:}
	(a) 36 randomly selected test samples.
	(b) The same samples with a central $5 \times 4$ block masked (47.6\% of pixels missing, shown in orange).
	(c)–(e) Reconstructions of (b) using the trained SRBM at 1, 500, and 1000 epochs, respectively, obtained by conditional LSB sampling.
	(f) The ratio of incorrect pixels per reconstruction as a function of training epoch.
	Each data point represents the mean over 10 independent runs with different random initializations of $\mathbf{u}$; error bars indicate the standard deviation.
	}
	\label{fig:BAS_Reconstruction}
\end{figure*}
We first use a dataset sampled from the exact Boltzmann distribution of a 3-spin model \cite{p-spin1,p-spin2}. 
The 3-spin model is an Ising system with random three-body interactions and has a highly rugged energy landscape. 
Consequently, the training data, drawn from its Boltzmann distribution, exhibit a nontrivial overlap distribution characterizing spin-glass systems (see the Supplemental Material~\cite{sp}). 
To allow exact computation of the cost function $D_{\rm KL}(P_D||Q_\beta)$ and enable quantitative comparison with conventional methods, we restricted $N_v$ to $10$. 
We prepared 10 independent instances to ensure statistically reliable performance evaluation.

Figure \ref{fig:3spin} compares the learning performance of four approaches: 
FBMs, RBMs and SRBMs trained with SAL (see Appendix\ \ref{sec:DetailsOfSAL}), 
and RBMs with $\beta = 1$ trained via contrastive divergence-$100$ (CD-$100$) \cite{RBM_guide}. 
Here, CD-100 means that the initial state is set to one of the data vectors and that 100 Monte Carlo steps of blocked Gibbs sampling (MCMC) are performed. 
The choice of 100 was made to match LSB sampling steps, $M=100$ (see the Supplemental Material~\cite{sp} for details of parameter setting). 
FBMs are fully connected Boltzmann machines without hidden variables ($N_h=0$) [see Fig.\ \ref{fig:1}(a)]. 
The number of hidden variables $N_h$ for both RBMs and SRBMs  [see Fig.\ \ref{fig:1}(b)-(c)] was set to $5$. 
All the four approaches allow parallel sampling.

Among the four models, FBMs exhibited the poorest performance [highest $D_{\rm KL}(P_D||Q_\beta)$]. 
This is natural because the 3-spin model includes non-pairwise (three-body) interactions that cannot be captured by FBMs, which rely solely on pairwise couplings. 
Introducing hidden variables, RBMs enabled the representation of higher-order correlations, leading to superior performance compared to FBMs. 
The observation that RBMs trained using SAL exhibit performance comparable to, or even surpassing, RBMs trained with CD-100 within the standard error suggests that SAL constitutes an effective learning strategy. 
Furthermore, SRBMs consistently achieved lower values of the cost function than RBMs across epochs. 
The difference between SRBMs and RBMs remained statistically significant even when considering the standard error, confirming the robustness of SRBM's advantage over RBMs. 
We also confirmed that SRBMs trained by replacing CEM in SAL with other computationally light but inaccurate inverse temperature estimation methods (maximum log-pseudo-likelihood estimation \cite{beta_estimation,QuALe}) or by using contrastive divergence with damped mean-field iterations \cite{HintonSRBM} resulted in significantly lower performance compared to SRBMs trained with SAL (see the Supplemental Material~\cite{sp}). 
This suggests that learning SRBMs, enabled by SAL, provides a distinct advantage, particularly for modeling highly rugged energy landscapes as exemplified by spin-glass systems.

\begin{figure*}[htbp]
	\includegraphics[width=\textwidth]{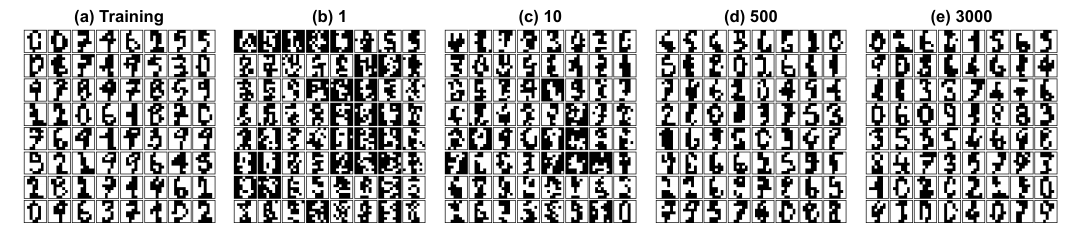}
	\caption{
	\textbf{Generative performance on the OptDigits dataset:} 
	(a) 64 randomly selected training samples.
	(b)–(e) 64 samples generated by the trained SRBM with LSB at 1, 10, 500, and 3000 training epochs, respectively.
	}
	\label{fig:OptDigitsGen}
\end{figure*}
\begin{figure*}[htbp]
	\includegraphics[width=1.0\textwidth]{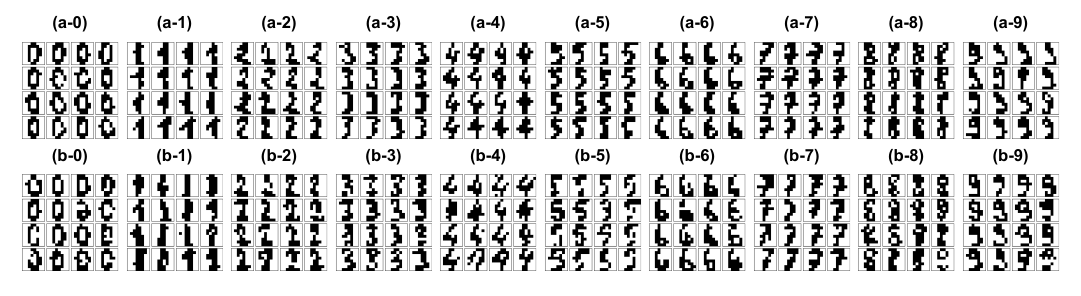}
	\caption{
	\textbf{Class-specified generation on the OptDigits dataset:} 
	(a-$n$) 16 randomly selected test samples for digit $n \in \{0,1,\cdots,8,9\}$.
	(b-$n$) 16 samples generated by conditional LSB sampling from the trained SRBM at 3000 epochs, conditioned on class label $n \in \{0,1,\cdots,8,9\}$.
	}
	\label{fig:OptDigitsCond}
\end{figure*}

\subsection{Generative and reconstruction performance on the BAS dataset}
We next consider the synthetic bars-and-stripes (BAS) dataset, consisting of binary images with horizontal and vertical stripe patterns (see Refs.\ [\onlinecite{BML3_Recognition}] and [\onlinecite{BAS2}]). 
We constructed a $7 \times 6$ BAS dataset with 192 distinct patterns (all possible patterns), split into $D = 96$ training and $D_{\rm Test} = 96$ test samples \cite{sp}. 
While exact computation of $D_{\rm KL}(P_D||Q_\beta)$ was possible for the 3-spin model with $N_v = 10$, 
it becomes challenging for the $7 \times 6$ BAS dataset ($N_v = 42$) due to the exponential growth of the state space. 
Therefore, 
here we focus on demonstrating the generative and reconstruction capabilities of SRBMs ($N_h = 21$) trained with SAL on the $7 \times 6$ BAS dataset, without computing $D_{\rm KL}(P_D||Q_\beta)$. 
Training settings are detailed in the Supplemental Material~\cite{sp}.

The generative capabilities are shown in Fig.\ \ref{fig:BAS_Generation}. 
The SRBM was trained via SAL on $D=96$ BAS patterns [panel (a) shows randomly selected 36 ones], 
and at various training epochs, 36 samples were generated by LSB from the trained SRBM [panels (b)–(f)].
As training progressed, the SRBM rapidly captured the essential features of both horizontal and vertical stripes, and at 6000 epochs [panel (f)]  generated only valid BAS patterns without misgeneration. 

Reconstruction experiment results shown in Fig.\ \ref{fig:BAS_Reconstruction} further demonstrate the robustness of SAL.
Here, we masked a central $5 \times 4$ block of each test image [Fig.\ \ref{fig:BAS_Reconstruction}(a)], resulting in a collapsed image with 47.6\% of pixels missing (shown in orange) [Fig.\ \ref{fig:BAS_Reconstruction}(b)].
By fixing the visible variables of the SRBM corresponding to the intact pixels in panel (b) to their observed values, we inferred the masked pixel values via conditional sampling [see Fig.\ \ref{fig:1}(f) and Appendix\ \ref{sec:ConditionalSampling}].
After 1000 epochs of learning, almost all patterns were successfully reconstructed by leveraging the learned distribution [Fig.\ \ref{fig:BAS_Reconstruction}(e)].
Figure \ref{fig:BAS_Reconstruction}(f) shows the fraction of incorrect pixels per reconstruction, evaluated across all $D_{\rm Test} = 96$ test samples.
This fraction rapidly dropped from 50.0\% (corresponding to random restoration) at 1 epoch to 0.5\% at 1000--3000 epochs, indicating near-perfect reconstruction across all test samples.
As training progressed beyond 3000 epochs, the fraction 
gradually increased to around 2\%, suggesting potential overfitting with excessive training.
Importantly, these results demonstrate that the SRBM trained with SAL learns the underlying distribution rather than memorizing individual samples, enabling robust and generalizable reconstruction even under substantial pixel loss.

\subsection{Generative and classification performance on the OptDigits dataset}
The OptDigits dataset consists of $8 \times 8$ images of handwritten digits and their class labels \cite{OptDigits}. 
We flattened and binarized the images, 
and represented the data class using one-hot encoding, 
resulting in $D = 3823$ training samples and $D_{\rm Test} = 1797$ test samples with $N_v = 74$ visible variables \cite{sp}.
Compared to the 3-spin model and BAS dataset, OptDigits has approximately 7.4-fold and 1.8-fold larger $N_v$, respectively, providing a more challenging benchmark. 
In the following, we demonstrate that the SRBM trained with SAL on OptDigits captures the data distribution for generative modeling and enables accurate image classification using the learned representation. 
Training settings are detailed in the Supplemental Material~\cite{sp}.

The generative performance is shown in Fig.\ \ref{fig:OptDigitsGen}.
The SRBM was trained via SAL on $D=3823$ digit images [panel (a) shows 64 randomly selected samples], and at various training epochs, 64 samples were generated by LSB sampling from the trained SRBM [panels (b)–(e)].
While samples at 1 epoch [panel (b)] exhibited little recognizable structure, subsequent epochs revealed a progressive emergence of digit-specific patterns consistent with the training data [panels (c)–(e)].
By 10 epochs [panel (c)], the generated images began to resemble the target digits, and by 500 and 3000 epochs [panels (d) and (e)], most samples captured the essential characteristics of handwritten digits and were visually recognizable.

\begin{figure}[t]
	\includegraphics[width=0.48\textwidth]{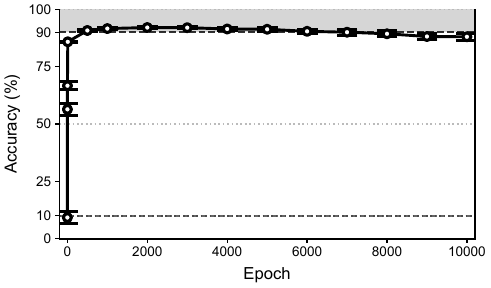}
	\caption{
	\textbf{Classification accuracy on the OptDigits dataset:} 
	The vertical axis shows the classification accuracy (\%), and the horizontal axis indicates training epochs. 
	Each data point represents the mean over 10 independent runs with different random initializations of $\mathbf{u}$; error bars indicate the standard deviation.
	The shaded region marks accuracy above 90\%, and the horizontal dashed lines indicate the 10\% baseline (random guessing) and the 90\% reference line.
	}
	\label{fig:OptDigitsAccuracy}
\end{figure}
We further evaluated the ability to perform conditional generation by fixing the class label $\mathbf{C}$ to the desired value and sampling the image variables $\mathbf{I}$ using LSB [see Fig.\ \ref{fig:1}(f), Appendix\ \ref{sec:ConditionalSampling}, and the Supplemental Material~\cite{sp}].
Figure\ \ref{fig:OptDigitsCond} illustrates the results.
For each $n \in \{0,1,\cdots,9\}$, panel (a-$n$) shows 16 randomly selected test images of digit $n$, 
and panel (b-$n$) shows 16 samples generated from the SRBM after 3000 training epochs by conditional LSB sampling with the digit class $n$.
Most of the generated images exhibited clear digit structures consistent with the specified condition, demonstrating that the SRBM successfully reproduced the overall data distribution and enabled class-conditional sample generation.

Finally, we evaluated the capability of the learned SRBM to perform image classification.
For each test sample $d\in\{1,\cdots,D_{\rm Test}\}$, the visible variables corresponding to the image $\mathbf{I}$ were fixed to their observed values $\mathbf{I}^{(d)}$, and conditional LSB sampling of $\mathbf{C}$ was performed to infer the class label $\mathbf{C}^{(d)}$ (see Appendix\ \ref{sec:ConditionalSampling} and the Supplemental Material~\cite{sp}).
The average label vector computed from conditionally sampled labels for each image $d$ was transformed into a one-hot representation $\tilde{\mathbf{C}}$ by setting $+1$ for the component with the highest value and $-1$ for all others.
Classification was deemed correct when this one-hot representation $\tilde{\mathbf{C}}$ matched the true class label $\mathbf{C}^{(d)}$.
Figure \ref{fig:OptDigitsAccuracy} shows that the classification accuracy improved substantially with training, rapidly reaching nearly $90\%$ at $500$ epochs from nearly 10\% (corresponding to random guessing in a 10-class problem) before training, underscoring the effectiveness of the proposed learning approach.

\section{Summary and Outlook}
In summary, we have proposed a fast and accurate non-MCMC Boltzmann sampler, Langevin simulated bifurcation (LSB), with highly parallelizable dynamics. 
We have also proposed conditional expectation matching (CEM), a method that resolves the inverse-temperature estimation problem inherent to fast non-MCMC samplers, not limited to LSB, when the underlying EBM exhibits suitable conditional independence structure. 
By combining these components, we have also developed sampler-adaptive learning (SAL), a framework that enables generative learning with non-MCMC samplers by adaptively matching the model temperature to that of the output distribution of a sampler. 
We demonstrated the effectiveness of the proposed methods (LSB, CEM, and SAL) on semi-restricted Boltzmann machines (SRBMs), a class of more expressive models that have been difficult to train using conventional methods due to the challenges in both sampling and learning.

Although SRBMs serve as a representative testbed, the proposed methods are not restricted to this specific model.
We have shown in Appendix~\ref{sec:SK} that LSB exhibits favorable sampling performance also on the Sherrington–Kirkpatrick model, suggesting that efficient sampling is possible for more complex EBMs. 
Furthermore, CEM and SAL rely only on structural properties such as conditional independence, and are therefore applicable to EBMs with exploitable conditional independence structures.
Exploring these applications and demonstrating practical implementations constitute important directions for future work. 
It is also of interest to investigate applications of non-MCMC sampling combined with CEM to neural quantum states \cite{BML_Quantum1}, where EBMs play a central role in representing complex quantum many-body systems.

Finally, we discuss several important remarks and open questions regarding LSB. 
As mentioned, the continuous limit of LSB resembles Langevin Monte Carlo \cite{LMC1,LMC2}, whose output distribution is guaranteed to be the Boltzmann distribution.
However, the theoretical basis for why LSB can well approximate the discrete Boltzmann distribution remains unclear, representing an important avenue for future research. 
Furthermore, developing systematic or automated strategies for hyperparameter selection in LSB is a topic that warrants further investigation.
While this work has mainly addressed the parallelizability of samplers, elucidating their relaxation properties is also an important subject for future research.

\section*{Author contributions}
K.K. conceived and developed LSB, CEM, and SAL; 
designed and performed all numerical experiments; 
analyzed the data; 
and drafted the manuscript. 
H.G. suggested the exploration of SB-based sampling; 
discussed the results with K.K.; 
improved the manuscript; 
and supervised the project. 

\section*{Competing interests}
K.K. and H.G. are inventors in Japanese patent applications related to this work, filed by Toshiba Corporation on 2 October 2025 under application numbers P2025-166272 and P2025-166278.
The authors declare no other competing interests.

\appendix{}
\section{Sampling Performance on SK Model}
\label{sec:SK}
\begin{figure*}[htbp]
	\includegraphics[width=1.0\textwidth]{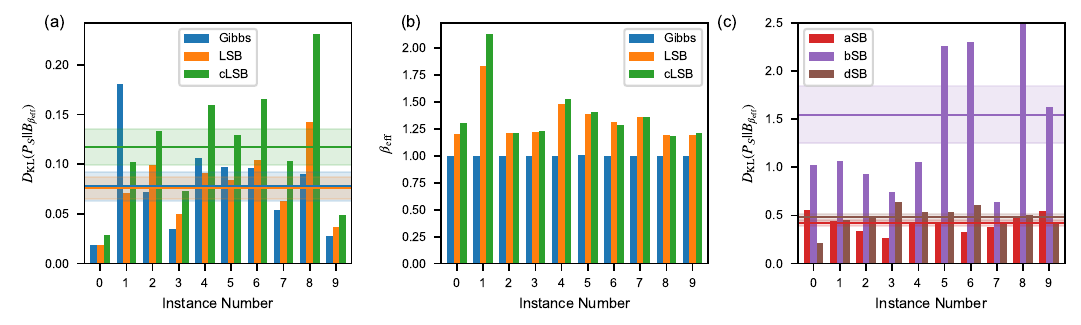}
	\caption{
	\textbf{Sampling accuracy $D_{\rm KL}(P_S|| B_{\beta_{\rm eff}})$ and effective temperature $\beta_{\rm eff}$ across ten SK model instances for various sampling methods:} 
	The horizontal axis of all panels indicates the instance number.
	Panel (a) shows $D_{\rm KL}(P_S|| B_{\beta_{\rm eff}})$ for LSB, Gibbs, and cLSB.
	Panel (b) shows $\beta_{\rm eff}$ estimated by minimizing the KL divergence corresponding to panel (a).
	Panel (c) shows $D_{\rm KL}(P_S|| B_{\beta_{\rm eff}})$ for aSB, bSB, and dSB.
	In panel (a) and (c), 
	horizontal lines and shaded regions indicate the means $D_{\rm KL}(P_S|| B_{\beta_{\rm eff}})$ and their standard errors.
	Parameter settings in these sampling experiments were as follows:
	The number of iteration steps for simulated birfurcation related methods and the number of Monte Carlo steps for Gibbs sampling were both set to $M=100$. 
	For LSB and cLSB, the discretization step size $\Delta$ was set to 1, and $\sigma$ was optimized from the candidate set $\sigma^{-2}\in \{0.5,0.6\cdots,1.9,2.0 \}$ to maximize sampling accuracy.
	For aSB, bSB, and dSB, parameters were set according to the notation in Ref.\ [\onlinecite{bSBdSB}]: $dt = 1.0$, $a_0 = 1.0$, $a(t) = 0$ for all methods, and $c_0$ was set to 1.0 for aSB, $\sqrt{\sum_{j=1}^N x_j^2/[\sum_{i=1}^N (\sum_{j=1}^NJ_{ij}x_j)^2]}$ for bSB, and $\sqrt{N/\{\sum_{i=1}^N [\sum_{j=1}^NJ_{ij}{\rm sgn}(x_j)]^2\}}$ for dSB.	
	}
	\label{fig:SK}
\end{figure*}

In this section, we study the sampling performance on the Sherrington–Kirkpatrick (SK) model:
\begin{align}
    E_{\rm{SK}}(\mathbf{s}) &= -\sum_{i=1}^{N}\sum_{j>i}^{N_v} J_{ij}s_{i}s_{j}, \\
    J_{ij} &\sim \mathcal{N}\left(0, \frac{\zeta}{\sqrt{N}}\right)\quad \text{for}\ 1\leq i<j\leq N,
\end{align}
where $E_{\rm{SK}}(\mathbf{s})$ denotes the global energy function of the SK model (the 2-spin model) \cite{p-spin1,p-spin2}, 
and $J_{ij}$ represents the pairwise interaction coefficients drawn from a Gaussian distribution $\mathcal{N}\left(0, \zeta/\sqrt{N}\right)$ with mean 0 and standard deviation $\zeta/\sqrt{N}$. 
The SK model is a fully connected spin-glass system, making the sampling task substantially more complex than the SRBM case discussed in Sec.\ \ref{sec:LSB}.
We executed sampling tasks for ten random instances, each characterized by a distinct set of $J_{ij}$ values. 
The parameter $\zeta$ was set to a sufficiently large value of 2 to yield a nontrivial spin-glass phase at $\beta=1$ \cite{p-spin1,p-spin2}. 
We sampled $D=9600$ Ising vectors $\mathbf{s}$ with $N=15$ using various sampling methods.

Figure \ref{fig:SK}(a) indicates that the sampling accuracy of LSB and that of Gibbs sampling with $\beta=1$ are comparable within the standard error, similarly to the results in Sec.\ \ref{sec:LSB}.
Here, cLSB refers to a variant of LSB in which the discretization procedure [Eq.~(\ref{eq:discretization})] is replaced by a clipping rule of bSB and dSB \cite{bSBdSB}:
\begin{align}
x_i \rightarrow
\begin{cases}
1 & \text{for } x_i \geq 1, \\
x_i & \text{for } -1<x_i<1, \\
-1 & \text{for } x_i \leq -1. \\
\end{cases}
\end{align}
Despite the relatively high sampling accuracy of cLSB, LSB achieves better sampling performance for all instances. 
This observation indicates that applying discretization at each iteration
plays a key role in enhancing the sampling accuracy.

Figure\ \ref{fig:SK}(b) shows the inverse temperature, obtained by minimizing the KL divergence, corresponding to sampling results of panel (a).
It was confirmed that the inverse temperature of the output distributions from LSB and cLSB also depends on the instance, similarly to what was observed in Sec.\ \ref{sec:LSB}.

Figure~\ref{fig:SK}(c) shows the sampling accuracy of conventional simulated bifurcation algorithms:
aSB, bSB, and dSB \cite{aSB,bSBdSB}, which, unlike LSB, do not incorporate stochastic initialization
or discretization applied at each iteration.
Although we tuned the parameters to achieve the best possible accuracy (see the caption of Fig.\ \ref{fig:SK} for details of parameters), the sampling accuracy of these conventional simulated bifurcations are clearly inferior to that of Gibbs sampling, LSB, and cLSB.
Among them, bSB exhibits the poorest performance [highest $D_{\rm KL}(P_S|| B_{\beta_{\rm eff}})$].
Although not shown in the figure, we have found that using variants such as HbSB and HdSB \cite{HbSBHdSB}, which are modified versions of bSB and dSB, can improve the sampling accuracy.
However, even with these improvements, the performance has not surpassed that of LSB and cLSB. 
These results indicate that the proposed enhancement strategy,
namely the application of discretization and stochastic initialization
at each iteration, plays a meaningful role in improving the sampling accuracy.

\section{Additional Computational Characteristics of LSB and CEM}
\label{sec:additional_computational}
In this section, we present additional analyses
to further characterize the computational properties of LSB and CEM
using the same benchmark problems as in Sec.\ \ref{sec:LSB} and Sec.\ \ref{sec:CEM}.
These analyses complement the results presented in the main text
and provide further insight into the practical characteristics
of the proposed sampling approach (LSB) and inverse-temperature estimation method (CEM).

\subsection{Results for Larger Systems}
\label{sec:larger_system}
To assess whether the qualitative behavior of LSB and CEM observed in
Sec.\ \ref{sec:LSB} and Sec.\ \ref{sec:CEM} persists when the system size
is moderately increased, 
we performed additional simulations for larger SRBMs
with $N_v=12$ and $N_h=6$, corresponding to a 20\% increase
over the original setting.
All other aspects of the experimental setup were kept the same as in Sec.\ \ref{sec:LSB} and Sec.\ \ref{sec:CEM}.

The results, shown in Fig.~\ref{fig:LSB_accuracy_CEM_larger},
indicate that the same qualitative trends observed in the smaller systems
are also present in this moderately larger setting.
In particular, the relative sampling performance of LSB
and the behavior of the effective inverse temperature
remain consistent with those reported in main text.
While this analysis does not constitute a systematic study of system-size scaling,
it provides empirical evidence that the observed behavior is not specific to the smallest problem instances considered.
\begin{figure}[t]
	\centering
	\includegraphics[width=0.5\textwidth]{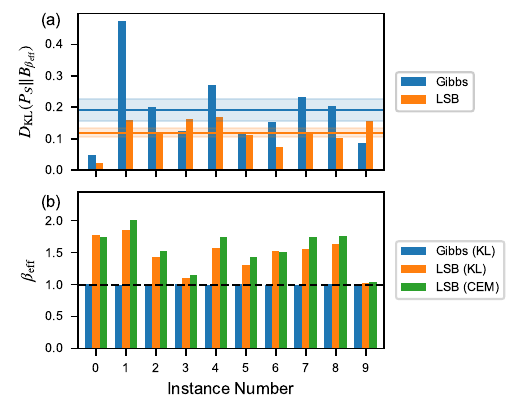}
	\caption{
	\textbf{Sampling accuracy and effective inverse temperature for LSB and Gibbs sampling:}
	(a) Sampling accuracy measured by the KL divergence
	$D_{\mathrm{KL}}(P_S \| B_{\beta_{\rm eff}})$
	for LSB and Gibbs sampling across 10 random instances of an SRBM
	with $N_v=12$ and $N_h=6$.
	Across the 10 instances, LSB achieved lower
	$D_{\mathrm{KL}}(P_S \| B_{\beta_{\rm eff}})$ values than Gibbs sampling
	in 8 out of 10 cases.
	The mean $D_{\mathrm{KL}}(P_S \| B_{\beta_{\rm eff}})$
	($\pm$ standard error) was $0.19\pm0.04$ for Gibbs sampling
	and $0.12 \pm 0.01$ for LSB.
	(b) Corresponding effective inverse temperature $\beta_{\rm eff}$
	of the output distributions produced by each sampler.
	The mean signed relative error between the $\beta_{\rm eff}$ values
	estimated by CEM and those obtained by minimizing the KL divergence for the output distributions produced by LSB
	is approximately 5.7\%.
	}
	\label{fig:LSB_accuracy_CEM_larger}
\end{figure}

\subsection{Computational Cost}
\label{sec:time}
We evaluate the computational cost of LSB by comparing its runtime with that of conventional Gibbs sampling under a fair CPU-based setting. 
Timing measurements were performed on the original problem instances considered in Sec.~\ref{sec:LSB} and Sec.~\ref{sec:CEM} using a system equipped with AMD EPYC 7742 processors. 
For the runtime comparison, we used 16 CPU cores to generate 16 samples, and results were averaged over 10 independent instances.
LSB sampling was implemented using NumPy-based
matrix--vector operations. 
In contrast, Gibbs sampling was parallelized
using Python multiprocessing, with 16 independent Markov chains
assigned to individual CPU cores.

Under these conditions, the measured elapsed time was $1.79\pm0.10\ \mathrm{ms}$ for LSB and $75.74\pm 5.58\ \mathrm{ms}$
for Gibbs sampling, indicating a substantially longer runtime for Gibbs sampling.
This difference is consistent with the intrinsic update structures of the two methods:
Gibbs sampling requires sequential conditional updates,
whereas LSB updates all $N$ variables simultaneously via matrix--vector operations
that are naturally amenable to parallel execution.

In learning scenarios, LSB may additionally require inverse-temperature
estimation via CEM when the effective
output temperature of the sampler needs to be estimated.
In practice, the dominant computational cost of CEM arises from the
conditional sampling steps, which require running LSB on a reduced
problem of size $N_h < N$.
Accordingly, the runtime associated with these conditional sampling steps
is bounded by that of LSB sampling itself.
In the present experiment with 16 samples, 
the elapsed time required for the conditional sampling within CEM was $0.89\pm0.02\ \mathrm{ms}$.
Consequently, even when LSB sampling and CEM are executed sequentially, the total runtime remains substantially
smaller than that of Gibbs sampling.
Furthermore, the inverse-temperature estimation procedure in CEM
is naturally parallelizable with LSB sampling,
since both rely on independent sampling operations.
Therefore, when sufficient computational resources are available, the total elapsed time of LSB including CEM is upper-bounded by the slower of the two components, which in typical settings is LSB itself.

Finally, we briefly comment on the memory cost.
In the present implementation, LSB and standard Gibbs sampling
require comparable memory usage for sampling itself,
as both methods store the same set of model parameters
and state variables.
When inverse-temperature estimation via CEM is incorporated,
additional memory is required to perform LSB sampling within CEM.
However, this additional cost remains limited,
since the sampling involved in CEM is applied only to the hidden variables.
As a result, even when LSB sampling and CEM are combined in parallel,
the total memory cost remains within a small constant factor
(of less than approximately twice) of that of LSB or Gibbs sampling alone.

\section{Conditional sampling}
\label{sec:ConditionalSampling}
In CEM and applications such as, image reconstruction, conditional generation, and classification [see Fig.\ \ref{fig:1}(f)], 
conditional sampling, which fixes a subset of variables while sampling the others, is often required. 
This is relatively straightforward with sequential MCMC methods; however, its implementation in non-MCMC samplers that perform parallel updates requires careful consideration. 
To achieve conditional sampling, Ref.\ [\onlinecite{Gray-box}] employed approaches that impose strong biases on the fixed variables in a D-Wave's quantum annealer. 
However, because both the inverse temperature and sampling accuracy depend strongly on $\mathbf{u}$ in LSB sampling, such methods raise concerns about potential changes in output characteristics when applied to LSB-based sampling.
Therefore, we adopt an alternative strategy as described below.

For simplicity, we consider 
the SRBM setting where a subset of visible variables $\mathbf{v}_f$ is fixed. 
This formulation covers all cases of conditional sampling in this study, and fixing more general variables, including hidden ones, can be formulated similarly. 
In this scenario, the energy of SRBMs, up to an additive constant, takes the following form:
\begin{align}
	E(\mathbf{v}_v,\mathbf{h}|\mathbf{v}_f, \mathbf{u}) 
	&= -\frac{1}{2}\mathbf{v}_v^{\rm T} V_{vv} \mathbf{v}_v
	-\mathbf{v}_v^{\rm T} W_v \mathbf{h} \nonumber\\
	&-(\mathbf{b}_v^{\rm T}+\mathbf{v}_f^{\rm T} V_{fv})\mathbf{v}_v
	-(\mathbf{c}^{\rm T}+\mathbf{v}_f^{\rm T} W_f)\mathbf{h}. 
\end{align}
Here, 
$\mathbf{v}_v$ denotes the unfixed visible variables, 
$V_{\mu\rho}$ $(W_{\mu})$ is a weight matrix between $\mathbf{v}_\mu$ and $\mathbf{v}_{\rho}$ 
(between $\mathbf{v}_{\mu}$ and $\mathbf{h}$) for $\mu,\rho\in\{v,f\}$, 
$\mathbf{b}_v$ is a bias vector of $\mathbf{v}_v$. 
The energy $E(\mathbf{v}_v,\mathbf{h}|\mathbf{v}_f, \mathbf{u})$ 
can be regarded as that of a conditional Boltzmann machine, $E(\mathbf{s}'|\mathbf{u}')$, defined over $\mathbf{s}'=(\mathbf{v}_v^{\rm T},\mathbf{h}^{\rm T})^{\rm T}$ with parameter set $\mathbf{u}'=(J',\mathbf{f}')$. 
Here, 
$\mathbf{f}'=({\mathbf{b}'}^{\rm T},{\mathbf{c}'}^{\rm T})^{\rm T}$, 
$\mathbf{b}'=\mathbf{b}_v+(\mathbf{v}_f^{\rm T}V_{fv})^{\rm T}$, 
$\mathbf{c}'=\mathbf{c}+(\mathbf{v}_f^{\rm T}W_{f})^{\rm T}$, 
and
\begin{align}
	J' = 
	\begin{pmatrix}
		V_{vv}\ & W_v\\
		W_v^{\rm T} & O_h \\
	\end{pmatrix},
\end{align}
where $O_h$ is an $(N_h\times N_h)$-dimensional zero matrix.

By executing LSB sampling for $E(\mathbf{s}'|\mathbf{u}')$ using the same hyperparameters as those used for sampling $B_{\beta_{\rm eff}}(\mathbf{s}|\mathbf{u})$, we obtain a conditionally sampled Ising vector set $\{{\mathbf{s}'}^{(l)}|l=1,\cdots,L'\}$. 
For the sake of simplicity in this study, we assumed $L' = L$.
We expect the empirical distribution $P_{{S}'}(\mathbf{s}')=\frac{1}{L'}\sum_{l'=1}^{L'}\delta(\mathbf{s}'-{\mathbf{s}'}^{(l')})$
to approximate the Boltzmann distribution of the conditional Boltzmann machine $B_{\beta_{\rm eff}}(\mathbf{s}'|\mathbf{u}')$, 
which shares the same inverse temperature $\beta_{\rm eff}$ as $B_{\beta_{\rm eff}}(\mathbf{s}|\mathbf{u})$. 
This conjecture is supported by numerical results obtained under the same conditions as Fig.\ \ref{fig:LSB_accuracy_CEM}(b), 
in which we confirmed that 
$D_{\rm KL}[P_{S'}(\mathbf{s}')||B_{\beta_{\rm eff}}(\mathbf{s}'|\mathbf{u}')]\ll D_{\rm KL}[P_S(\mathbf{s})||B_{\beta_{\rm eff}}(\mathbf{s}|\mathbf{u})]$.

Note that $B_{\beta_{\rm eff}}(\mathbf{s}'|\mathbf{u}')$ corresponds to the conditional Boltzmann distribution of the original Boltzmann machine: 
\begin{align}
	A_{\beta_{\rm eff}}(\mathbf{s}'|\mathbf{v}_f,\mathbf{u})&
	=\frac{B_{\beta_{\rm eff}}(\mathbf{s}|\mathbf{u})}{\sum_{\mathbf{v}_v}\sum_{\mathbf{h}}B_{\beta_{\rm eff}}(\mathbf{s}|\mathbf{u})}. 
\end{align}
In CEM, 
we set $\mathbf{v}_f$ to $\mathbf{v}$ and $\mathbf{s}'$ to $\mathbf{h}$. 
We then compute sampling-based estimation of $\langle h_j \rangle_{\mathbf{r},\beta_{\rm eff}}$, 
\begin{align}
	\langle h_j \rangle_{\mathbf{r},C}:=\sum_{\mathbf{h}} h_j P_{S'}(\mathbf{h}), 
\end{align}
by conditional LSB sampling. 
This is compared to the analytical expression of $\langle h_j \rangle_{\mathbf{r},\beta_{\rm eff}}$ [Eq.\ (\ref{eq:hi})] to estimate $\beta_{\rm eff}$ (see Sec.\ \ref{sec:CEM}).

In image restoration, conditional generation, and classification, 
$\mathbf{v}_f$ are set to the intact image pixels, 
the target class label $\mathbf{C}$, and the image $\mathbf{I}$ of the test data, respectively.

\section{Learning rule of SAL
\label{sec:DetailsOfSAL}}
In energy-based generative modeling, the goal is to minimize the
KL divergence between the target distribution $P_D$ and $Q_{\beta}$ defined in Eq.\ (\ref{eq:DKL}), typically using gradient-based methods.
In SAL, where the inverse temperature $\beta$ is set to $\beta_{\rm eff}$, the inverse temperature of the output distribution produced by the sampler, the parameters of SRBMs are updated as follows:
\begin{widetext}
\begin{align}
	&V_{ij}(k+1) = V_{ij}(k)-\eta \left[\left\langle v_i v_j \right\rangle_{D}-\langle v_i v_j \rangle_{\beta_{\rm eff}}\right], 
	 \label{eq:V}
	\\
	&W_{ij}(k+1) = W_{ij}(k)-\eta \left\{\left\langle v_i  \tanh \left[\beta_{\rm eff}\left(c_{j}+\sum_{i=1}^{N_v}v_i W_{ij}\right)\right] \right\rangle_{D}
	-\langle v_i h_j \rangle_{\beta_{\rm eff}}\right\}, 
	\label{eq:W}
	\\
	&b_{i}(k+1) = b_{i}(k)-\eta \left[\left\langle v_i \right\rangle_{D}-\langle v_i \rangle_{\beta_{\rm eff}}\right], 
	 \label{eq:b}
	\\
	&c_{j}(k+1) = c_{j}(k)-\eta \left\{\left\langle \tanh \left[\beta_{\rm eff}\left(c_{j}+\sum_{i=1}^{N_v}v_i W_{ij}\right)\right]
 \right\rangle_{D}-\langle h_j \rangle_{\beta_{\rm eff}}\right\},
 	\label{eq:c}
\end{align}
\end{widetext}
where $k$ denotes the learning step, 
$\eta>0$ is the $\beta_{\rm eff}$-renormalized learning rate \cite{Gray-box}, 
$\langle\bullet\rangle_{D}$ indicates the expectation value under the data distribution $P_{D}(\mathbf{v})$, 
and $\langle\bullet\rangle_{\beta_{\rm eff}}$ indicates the expectation value under the Boltzmann distribution $B_{\beta_{\rm eff}}(\mathbf{s})$ with inverse temperature $\beta_{\rm eff}$. 
The terms defined by $\langle\bullet\rangle_{D}$ are referred to as the positive phase, whereas those defined by $\langle\bullet\rangle_{\beta_{\rm eff}}$ are referred to as the negative phase.
These update rules assume that variations in $\beta_{\rm eff}$ are small 
compared to changes in $\mathbf{u}$. 
The negative phase is computed using LSB as follows:
\begin{align}
	\langle\bullet\rangle_{\beta_{\rm eff}}\simeq \sum_{l=1}^{L} P_{S}(\mathbf{s})\bullet, 
\end{align}
where
$P_S(\mathbf{s})=\frac{1}{L}\sum_{l=1}^{L}\delta(\mathbf{s}-\mathbf{s}^{(l)})$ 
represents the empirical distribution of Ising vectors $\{\mathbf{s}^{(l)}|l=1,\cdots,L\}$ sampled by LSB. 
Thus, the negative phase does not require explicit information about $\beta_{\rm eff}$.
On the other hand, computing the positive phase associated with the hidden units [Eqs.\  (\ref{eq:W}) and (\ref{eq:c})] requires the effective inverse temperature $\beta_{\rm eff}$, which is estimated using CEM.

Setting the initial values of $V_{ij}$ to zero and updating only $W_{ij}$, $b_i$, and $c_j$ reduces the update rule to that of an RBM. 
Similarly, setting $N_h = 0$ and updating only $V_{ij}$ and $b_i$ reduces the update rule to that of an FBM. 
In the FBM case, only Eqs.\ (\ref{eq:V}) and (\ref{eq:b}) remain, and $\beta_{\rm eff}$ estimation is no longer required. 
The FBM version of SAL has already been reported as the Gray-box algorithm \cite{Gray-box}. 
That study reported that when hidden variables are present, computing the positive phase becomes difficult, and consequently they restricted their approach to FBMs, which can be trained without explicit knowledge of $\beta_{\rm eff}$. 
SAL overcomes this challenge by leveraging SRBM's (and RBM's) structure together with CEM for efficient $\beta_{\rm eff}$ estimation, enabling scalable learning without sacrificing expressive power.

We note that the same learning principle is applicable to EBMs with higher-order interactions in the visible units, yielding similarly simple update rules.  In such cases, only the update rules for the visible-unit coupling coefficients need to be modified.

\clearpage 
\onecolumngrid 
\begin{center} 
{\large\bfseries Supplemental Material for} 
\vspace{0.15cm} 

{\large\bfseries Training Energy-Based Models with Non-MCMC Samplers and Efficient Temperature Estimation } 
\vspace{0.6cm} 

Kentaro Kubo$^{1,*}$ and Hayato Goto$^{1,2}$ 
\vspace{0.2cm} 

$^{1}$\textit{Corporate Laboratory, Toshiba Corporation, Kawasaki, Kanagawa 212-8582, Japan} 

$^{2}$\textit{RIKEN Center for Quantum Computing (RQC), Wako, Saitama 351-0198, Japan} 
\vspace{0.2cm} 

(Dated: \today) 
\end{center} 
\vspace{0.5cm}
\setcounter{section}{0} 
\setcounter{figure}{0} 
\setcounter{table}{0} 
\setcounter{equation}{0} 
\renewcommand{\thesection}{S\arabic{section}} 
\renewcommand{\thefigure}{S\arabic{figure}} 
\renewcommand{\thetable}{S\arabic{table}} 
\renewcommand{\theequation}{S\arabic{equation}} 

\twocolumngrid
\section{Comparison with Other $\beta_{\rm eff}$ estimation methods}
\label{sec:CEMn_MLPL}
In Fig.\ \ref{fig:LSB_accuracy_CEM_MLPL}, we compare $\beta_{\rm eff}$ estimated by various methods for LSB output distribution shown in Sec.\ III of the main text.
The results of CEM and minimizing the KL divergence are identical to those in Fig.\ 2(b) of the main text. 

In CEM-$n$ approach, we consider $n$ distinct conditions $\{\mathbf{r}^{(m)}|m=1,\cdots,n\}$ and estimate $\beta_{\rm eff}$ using the modified expression: $\beta_{\rm eff}={\rm argmin}_{\beta>0} \sum_{m=1}^{n}F_{\mathbf{r}^{(m)}}(\beta)$. 
CEM introduced in Sec.\ IV of the main text is identical to CEM-1.
For CEM-$3$ and $5$, each component of the condition vectors $\{\mathbf{r}^{(m)}|m=1,\cdots,n\}$ was uniformly sampled in $\{-1,1\}$.
The mean signed relative error between them across 10 instances is approximately $3.6\pm1.3\%$, $3.9\pm1.3\%$, and $3.8\pm1.3\% $ for CEM-$n$ with $n=1, 3,$ and $5$, respectively.
These results indicate that CEM-$n$ ($n\in\{1,3,5\}$) exhibit comparable performance within the standard error. 
Furthermore, the estimation error of $\beta_{\rm eff}$ using CEM-$n$ tends to shift toward lower temperatures.

The MLPL in Fig.\ \ref{fig:LSB_accuracy_CEM_MLPL} refers to maximum log-pseudo-likelihood estimation \cite{beta_estimation,QuALe}, a method commonly regarded as efficient, though its accuracy has been questioned. 
In the results, we found that its signed relative error to KL was $-4.0\pm1.1\%$, which was comparable in absolute magnitude to that of CEM. However, MLPL exhibited a bias toward higher estimated temperatures, in contrast to CEM. 
Furthermore, when MLPL was used instead of CEM for inverse temperature estimation in SAL, the learning performance of SRBMs deteriorated significantly (see Sec.\ \ref{sec:det_3spin}). This deterioration can be attributed to the opposing bias tendencies in the inverse temperature estimates produced by MLPL and CEM.
\begin{figure}[htbp]
	\centering
	\includegraphics[width=0.5\textwidth]{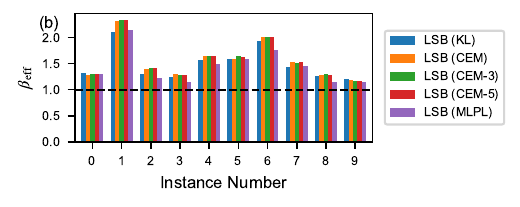}
	\caption{
	\textbf{Effective inverse temperature for LSB:}
	The effective inverse temperature $\beta_{\rm eff}$ for the output distribution produced by LSB. 
	Inverse temperature $\beta_{\rm eff}$ was estimated by minimizing the KL divergence, CEM-$n$ with $n=1, 3, 5$ where $\{\mathbf{r}^{(m)}|m=1,\dots,n\}$ were randomly generated Ising vectors, and MLPL. 
	}
	\label{fig:LSB_accuracy_CEM_MLPL}
\end{figure}

\section{Details of Experiments}
\label{sec:DetailsOfExperiments}
In this section, we describe the experimental settings in Sec.\ V of the main text. 

\subsection{Summary of Parameter Settings}
\begin{table}[htbp]
	\caption{Summary of parameters for each experiment.}
	\centering
	\begin{tabular}{ccccccccc}
		\hline
		\hline
		Data set & $N_{v}$ & $N_h$ &$N$& $D$ & $D_{\rm Test}$ & $L=L'$ & $M$ & $\eta$ \\
		\hline
		3-spin model & 10 & 5 & 15 & 9600 & N/A & 9600 & 100 & 0.05 \\ 
		BAS & 42 & 21 & 63 & 96 & 96 & 96 & 250 &0.001 \\
		OptDigits & 64+10 & 37 & 111 & 3823 & 1797 & 320 & 500 & 0.001\\
		\hline 
		\hline
	\end{tabular}
	\label{tab:params}
\end{table}
Most parameter values used in the experiments in Sec.\ V of the main text are summarized in Table~\ref{tab:params}. 
Common settings and additional details of sampling parameters are described below.

All training was performed with momentum $\alpha = 0.5$ and $L_2$ regularization $\lambda = 10^{-5}$ applied to $V_{ij}$ and $W_{ij}$. 
The initial values of nonzero $J_{ij}$ were sampled from $\mathcal{N}(0, 10^{-4})$ 
and the initial values of $f_i$ were set to zero. 

For LSB sampling, the hyperparameter $\Delta$ was fixed to 1 and 
$\sigma$ was selected by empirically testing several candidate values and choosing the one that gave good learning performance. 
For the 3-spin model, $\sigma$ was optimized for each instance across the candidate set $\sigma^{-2}\in\{0.5, 0.6, \dots, 1.9, 2.0\}$. 
For the BAS and OptDigits datasets, $\sigma$ was set as $\sigma^{-2}=1.0$ for the BAS and $\sigma^{-2}=3.5$ for the OptDigits.
The selected values were kept fixed during each training. 
For comparison, CD-100 was also used for the 3-spin model with $\beta=1$ to train RBMs. 

In CEM, the condition vector $\mathbf{r}$ was set to a data vector $\mathbf{v}^{(d)}$ randomly selected from the training dataset.

For OptDigits only, mini-batch learning was employed with a batch size of 382, resulting in ten mini-batches per epoch. 
Before splitting, the entire training dataset was randomly shuffled to ensure that each mini-batch contained approximately equal proportions of samples from each class.

\subsection{Dataset Preparation}
\textbf{3-spin model:} 
Data were generated according to the exact Boltzmann distribution:
\begin{align}
	B_{\rm{3spin}}(\mathbf{v})&=\frac{e^{-E_{\rm{3spin}}(\mathbf{v})}}{\sum_{\mathbf{v}}e^{-E_{\rm{3spin}}(\mathbf{v})}}, \\
	E_{\rm{3spin}}(\mathbf{v})&=-\sum_{i=1}^{N_v}\sum_{j>i}^{N_v}\sum_{k>j}^{N_v}T_{ijk}v_{i}v_{j}v_{k}, \\
	T_{ijk}&\sim\mathcal{N}\left(0,\frac{\sqrt{3}\zeta}{N_v}\right)\ {\rm for}\ 1\leq i<j<k\leq N_v,
\end{align}
where $E_{\rm{3spin}}(\mathbf{v})$ denotes the global energy function of the 3-spin model \cite{p-spin1,p-spin2}, 
and $T_{ijk}$ represents the three-body interaction coefficients drawn from a Gaussian distribution $\mathcal{N}\!\left(0, \sqrt{3}\zeta/N_v\right)$ with mean 0 and standard deviation $\sqrt{3}\zeta/N_v$. 
Ten independent instances were generated, each characterized by a distinct set of $T_{ijk}$ values. 
Each dataset consisted of $D=9600$ samples with $N_v=10$, obtained via exact Boltzmann sampling by enumerating all visible configurations $\mathbf{v}$ and drawing samples based on exhaustively computed Boltzmann weights $B_{\rm{3spin}}(\mathbf{v})$.
We set $\zeta$ to a sufficiently large value of 2, which is enough to yield a nontrivial spin-glass phase in the thermodynamic limit \cite{p-spin1,p-spin2}. 
The overlap distribution of the generated samples was verified to exhibit spin-glass-like features (see Sec.\ \ref{sec:det_3spin}).
\begin{figure*}[htbp]
	\includegraphics[width=\textwidth]{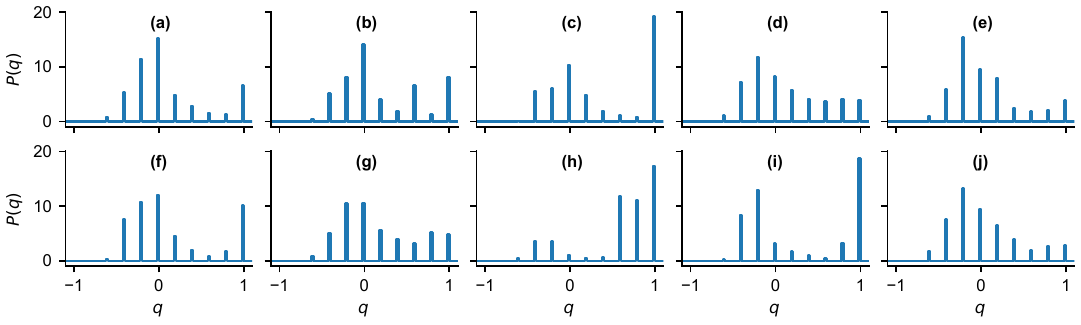}
	\caption{
	\textbf{Overlap distributions for all instances:}  
	Each panel shows the overlap distribution $P(q)$ for one of the 10 dataset instances. 
	All instances exhibit nontrivial spin-glass features.
	}
	\label{fig:overlap}
\end{figure*}

\textbf{BAS:} 
The bars-and-stripes (BAS) dataset consists of binary images generated by assigning each row or column to either black ($-1$) or white ($+1$) at random\ \cite{BML3_Recognition,BAS2}. 
We constructed a BAS dataset of $7 \times 6$ binary images, containing 192 mutually distinct patterns (all possible patterns), each with $N_v = 42$ binary variables.
The images were randomly shuffled and split into training and test sets of equal size ($D = D_{\rm Test} = 96$).

\textbf{OptDigits:} 
The Optical Recognition of Handwritten Digits (OptDigits) dataset was obtained from the UCI Machine Learning Repository\ \cite{OptDigits}. 
Each original sample consists of a four-bit grayscale image of $8 \times 8$ pixels and a categorical label indicating one of ten digit classes (0--9). 
For our experiments, each image was flattened into a 64-dimensional image vector $\mathbf{I}$, and the class label was converted to a 10-dimensional one-hot encoded vector $\mathbf{C}$. 
Specifically, for a sample belonging to class $j\in\{0,\cdots,9\}$, the $j$-th component of $\mathbf{C}$ was set to $+1$, while all other components were set to $-1$. 
Pixel values were binarized by thresholding at the midpoint of the original four-bit gray scale range, mapping them to $\{-1,+1\}$, thus each image vector $\mathbf{I}$ was converted into a 64-dimensional Ising vector. 
We employed combined representation $\mathbf{v} = (\mathbf{I}^{\rm T}, \mathbf{C}^{\rm T})^{\rm T}$ as data vector with $N_v = 74$ variable. 
The final dataset contained $D = 3823$ training samples and $D_{\rm Test} = 1797$ test samples, with approximately equal proportions across classes.

\section{Additional Information on the 3-spin model}
\label{sec:det_3spin}
\subsection{Overlap Distribution}
We analyze the overlap distribution of the 3-spin model dataset discussed in Sec.\ V A of the main text.
For a given instance $\{\mathbf{v}^{(d)}|d=1,\cdots,D\}$, we consider the overlap between $\mathbf{v}^{(\mu)}$ and $\mathbf{v}^{(\rho)}$:
\begin{align}
	q^{(\mu, \rho)}=\frac{1}{N_v}\sum_{i=1}^{N_v} v_i^{(\mu)}v_i^{(\rho)},
\end{align}
and construct the corresponding histogram across all samples $P(q)$\cite{p-spin1,p-spin2}. 
All instances exhibit nontrivial overlap distributions that are neither paramagnetic, characterized by a single peak centered at $q=0$, nor ferromagnetic or antiferromagnetic, both characterized by two peaks located near $q=\pm1$, as shown in Fig.\ \ref{fig:overlap}.
These observations indicate that the dataset reflects a genuine spin-glass regime and possesses nontrivial structures.

\subsection{Sampling Accuracy and $\beta_{\rm eff}$ During Training}
This subsection tracks the evolution of sampling accuracy, measured by $D_{\rm KL}(P_S \Vert B_{\beta_{\rm eff}})$, and the effective inverse temperature $\beta_{\rm eff}$ throughout the learning process described in Sec.~V A of the main text.

\begin{figure}[htbp]
	\includegraphics[width=0.48\textwidth]{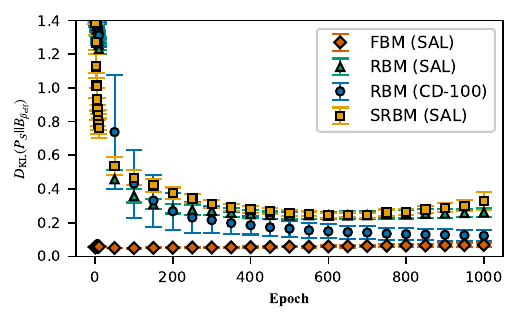}
	\caption{
	\textbf{Evolution of sampling accuracy $D_{\rm KL}(P_{S} || B_{\beta_{\rm eff}})$ during training on the 3-spin model:}  
	The horizontal axis shows the training epoch, and the vertical axis shows the KL divergence between the sampled distribution $P_{S}$ and the Boltzmann distribution $B_{\beta_{\rm eff}}$ at the estimated effective inverse temperature $\beta_{\rm eff}$.
	}
	\label{fig:3spin_sampling_accuracy}
\end{figure}
Figure \ref{fig:3spin_sampling_accuracy} shows the evolution of sampling accuracy, measured by $D_{\rm KL}(P_{S} || B_{\beta_{\rm eff}})$, during training on the 3-spin model discussed in Sec.~V A of the main text. 
LSB in FBMs achieved the smallest $D_{\rm KL}(P_{S} || B_{\beta_{\rm eff}})$, which is reasonable because FBMs use $N=10$ visible units, whereas RBMs and SRBMs use $N=15$, making sampling inherently more challenging for the latter. 
Nevertheless, it is noteworthy that the sampling accuracy achieved by LSB in FBMs surpassed that of CD-100 in RBMs.
The sampling accuracy of LSB in SRBMs and RBMs was comparable to CD-100 in RBMs during the early stages. 
Although the former became slightly inferior after about 350 epochs, the overall training performance remained comparable to or even surpassed that of the latter, as detailed in Sec.~V A of the main text.
These results highlight the effectiveness of LSB as a parallel sampler for Boltzmann machine learning.

\begin{figure*}[htbp]
	\includegraphics[width=\textwidth]{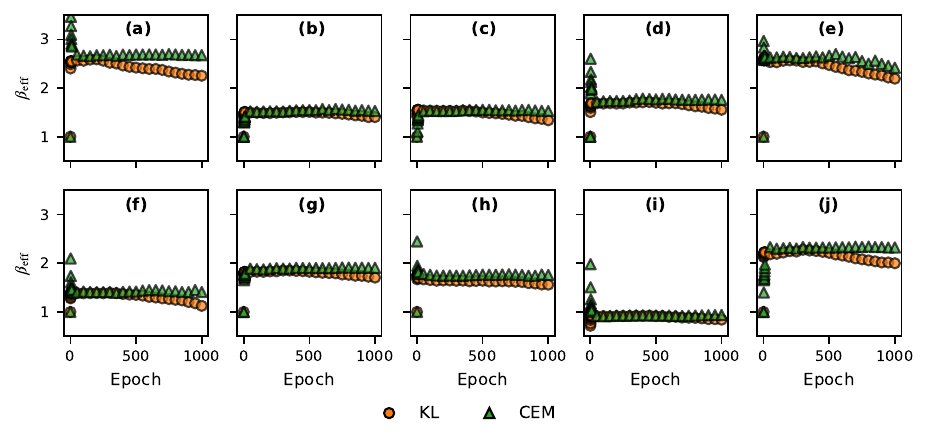}
	\caption{
	\textbf{Evolution of the effective inverse temperature $\beta_{\rm eff}$ during training on the 3-spin model:} 
	Each panel (a)–(j) shows the trajectory of $\beta_{\rm eff}$ estimated by the KL-based method and CEM for individual instances. 
	The horizontal axis represents the training epoch, and the vertical axis indicates the estimated value of $\beta_{\rm eff}$.
	}
	\label{fig:3spin_sampling_beta}
\end{figure*}
Figure~\ref{fig:3spin_sampling_beta} illustrates how the effective inverse temperature $\beta_{\rm eff}$ of SAL-trained SRBMs evolves throughout the training process 
in Sec.\ V A of the main text.
For each instance (panels a–j), the trajectories of $\beta_{\rm eff}$ estimated by KL-based method and CEM are compared across training epochs. 
Although there are a few exceptional instances, such as (a), the CEM estimators generally provide stable and reliable estimates of $\beta_{\rm eff}$ throughout training, closely matching the KL-based estimator.
Although not shown, a similar trend was observed for RBMs as well.
These results underscore the robustness of CEM for inverse temperature estimation in practical Boltzmann machine learning.

\subsection{Comparison with other learning methods}
\label{ap:ComparisonWithOtherLearningMethods}
Figure \ref{fig:3spin_others} shows the learning performance of SRBMs trained with SAL to RBMs and SRBMs trained using alternative methods not discussed in Sec.\ V A of the main text.

First, the performance of RBMs trained with CD-1 was markedly inferior to that of Blocked Gibbs (BG) sampling with 100 iterations, which yields performance nearly equivalent to CD-100 in in Sec.\ V A of the main text. 
In contrast, CD-10 achieved performance comparable to BG, highlighting the rapid convergence of the CD method. 
However, even when RBMs were successfully trained using CD-10 or BG their performance did not reach that of SRBMs trained with SAL. 

Replacing CEM in SAL with the MLPL (maximum log-pseudo-likelihood) $\beta_{\rm eff}$ estimator \cite{beta_estimation,QuALe} resulted in SRBM performance nearly identical to that of RBMs trained with CD-10 or BG. 
This performance degradation may be attributed to the opposing bias tendencies of CEM and MLPL (see Sec.\ \ref{sec:CEMn_MLPL}).

Because visible variables are not independent, CD is theoretically inappropriate in SRBMs. 
In our experiments, CD-1 for learning SRBMs led to severe degradation, with $D_{\rm KL}(P_D||Q_\beta)$ increasing substantially as learning progressed.
Although not shown in the figure, SRBMs trained with CD-10 and CD-100 exhibited even poorer and highly unstable performance. 
Previous reports have suggested that CD combined with damped mean-field iterations (DMFI) for visible variables could be effective for SRBMs \cite{HintonSRBM,SRBM3}. 
However, this method also achieved even lower performance than RBMs trained with CD-1, highlighting the limitations of CD-based methods for SRBMs. 

These findings suggest that SAL provides an effective framework for enabling SRBMs to 
better exploit their enhanced representational capacity, 
leading to performance improvements beyond those achieved by conventional Boltzmann machine learning. 
\begin{figure}[htbp]
	\includegraphics[width=0.48\textwidth]{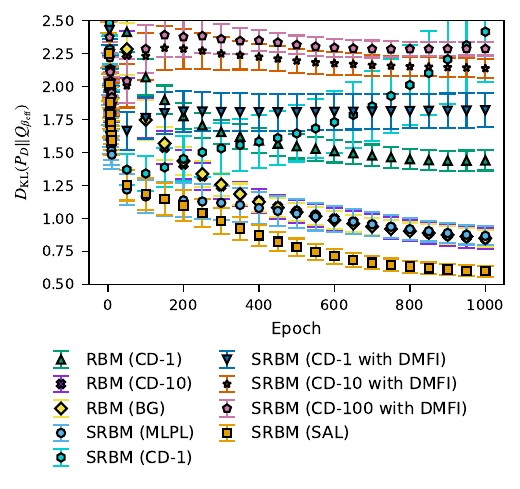}
	\caption{
	\textbf{Further comparison of learning performance on the 3-spin model:} 
	The vertical axis shows the cost function $D_{\rm KL}(P_D||Q_\beta)$, and the horizontal axis indicates training epochs. 
	Nine models are compared: 
	RBMs trained using CD-1, CD-10, and Blocked Gibbs (BG) sampling with 100 Monte Carlo steps; 
	SRBMs trained using SAL in which CEM was replaced with MLPL \cite{beta_estimation,QuALe}; 
	SRBMs trained using CD-1; 
	SRBMs trained using CD-1, CD-10, and CD-100 with damped mean-field iterations (DMFI) \cite{HintonSRBM}; 
	and SRBMs trained using SAL. 
	In the DMFI procedure, we used 5 iterations with a damping factor of 0.5.
	The number of hidden variables $N_h$ for both RBMs and SRBMs was set to 5 as in Sec.\ V A of the main text. 
	Each point represents the mean over the same 10 independent instances used for training in Sec.\ V A of the main text, and error bars denote the standard error. 
	}
	\label{fig:3spin_others}
\end{figure}

\end{document}